%% file: main.tex
\documentclass[10pt,twocolumn,letterpaper]{article}
\usepackage[T1]{fontenc}
\usepackage{times}
\usepackage[letterpaper,left=.78in,right=.78in,top=.82in,bottom=.9in,footskip=.42in]{geometry}
\usepackage{amsmath,amssymb,graphicx,booktabs,caption,float,placeins}
\usepackage{algorithm,algpseudocode}
\usepackage[numbers,sort&compress]{natbib}
\usepackage{xcolor}
\definecolor{linkblue}{HTML}{285A8A}
\usepackage[breaklinks,colorlinks,allcolors=linkblue]{hyperref}
\hypersetup{pdftitle={Shift by Analogy for Compositional Zero-Shot Learning},pdfauthor={Zanwu Liu, Bo Li, Guanglin Niu}}
\makeatletter
\renewcommand{\section}{\@startsection{section}{1}{\z@}{-2.2ex plus -.5ex minus -.2ex}{.85ex plus .15ex}{\normalfont\large\bfseries}}
\renewcommand{\subsection}{\@startsection{subsection}{2}{\z@}{-1.8ex plus -.4ex minus -.2ex}{.6ex plus .1ex}{\normalfont\normalsize\bfseries}}
\makeatother
\begin{document}
\raggedbottom
\twocolumn[{\centering
{\LARGE\bfseries Shift by Analogy for Compositional Zero-Shot Learning\par}
\vspace{11pt}
{\large Zanwu Liu\qquad Bo Li\qquad Guanglin Niu\textsuperscript{*}\par}
\vspace{4pt}
School of Artificial Intelligence, Beihang University, Beijing, China\par
{\small \textsuperscript{*}Corresponding author: \href{mailto:beihangngl@buaa.edu.cn}{beihangngl@buaa.edu.cn}\par}
\vspace{2pt}
{\small \href{https://github.com/ashmentlzw/Shift-by-Analogy}{github.com/ashmentlzw/Shift-by-Analogy}\par}
\vspace{14pt}
}]
\input{sections/0_abstract}
\input{sections/1_introduction}

\input{figures/method/method}
\input{sections/2_related_work}

\input{sections/3_method}

\FloatBarrier
\input{sections/4_experiments}

\FloatBarrier
\input{sections/5_conclusion}
\clearpage
\section*{LLM Usage Statement}
ChatGPT (Codex) assisted with polishing and typesetting the manuscript.
{\fontsize{9}{10}\selectfont
\setlength{\bibsep}{0pt}
\nocite{liu2026vpfa,liu2026masp}
\bibliographystyle{ieeenat_fullname-shortdash}
\bibliography{references}
}
\clearpage
\appendix
\setcounter{table}{0}
\setcounter{figure}{0}
\setcounter{equation}{0}
\renewcommand{\thetable}{S\arabic{table}}
\renewcommand{\thefigure}{S\arabic{figure}}
\renewcommand{\theequation}{S\arabic{equation}}
\renewcommand{\theHtable}{supp.\arabic{table}}
\renewcommand{\theHfigure}{supp.\arabic{figure}}
\renewcommand{\theHequation}{supp.\arabic{equation}}
\twocolumn[{
\begin{center}\Large\bfseries Supplementary Material\end{center}
\section{Additional Comparisons}\label{app:recent}
\begin{minipage}{\textwidth}\captionsetup{type=table}
\input{tables/supplement/complete/comparison}
\end{minipage}\vspace{10pt}
}]
\input{sections/6_appendix}
\end{document}

%% file: sections/0_abstract.tex
\begin{abstract}
Compositional zero-shot learning recognizes attribute-object pairs absent from training. An unseen pair has a semantic description but lacks an observed visual reference. We propose \textbf{Shift by Analogy (SBA)} to make unseen compositions visually seen through a \emph{shift-then-match} process. Its core \emph{add-subtract-add} rule transfers an attribute change between two seen image-text residuals on a source object to a reference residual on the target object. This shifts the target's own text feature toward its visual center; learned refinement produces a proxy for matching query images. During training, each target is excluded from its supports, and recognition is learned using references constructed from other pairs. An analogy library and same-object attention extend this construction across candidates, while joint learning integrates proxy matching with semantic and primitive evidence. Visual-center analysis confirms substantially improved correspondence for unseen compositions. Experiments across three standard benchmarks show that SBA consistently outperforms strong CZSL baselines under both closed- and open-world evaluation.
\end{abstract}

%% file: sections/1_introduction.tex
\section{Introduction}
\label{sec:intro}
Compositional zero-shot learning (CZSL) recognizes attribute-object pairs absent from training by recombining known primitives~\cite{misra2017red,nagarajan2018operators,purushwalkam2019modular}. Existing approaches learn transferable attribute and object representations and combine them to recognize new pairs. Pretrained vision-language models such as CLIP~\cite{radford2021clip} represent candidate pairs through language in a shared image-text space. Building on this foundation, recent methods learn compositional prompts~\cite{nayak2023csp,lu2023dfsp} and coordinate primitive and composition representations~\cite{huang2024troika,zhang2026eva}, improving the transfer of known concepts to new combinations.

\suppressfloats[t]
\input{figures/teaser/teaser}

Visual proxies~\cite{zhang2025proxy} and compositional flows~\cite{he2026flowcomposer} further model visual references and transformations. The central challenge is to establish a visual reference for a composition whose images are absent from training. Its name specifies which primitives are combined, while its appearance depends on how the attribute interacts with the object. Seen pairs obtain this correspondence from images; unseen pairs must acquire it through transfer. We therefore leverage observed relations among seen pairs to turn each missing composition into an available visual reference.

Our central idea is to \emph{make unseen compositions visually seen} by constructing the references they lack. Figure~\ref{fig:motivation}(a) illustrates the missing sliced-bread reference: training provides fresh tomato, sliced tomato, and fresh bread. The first two reveal the fresh-to-sliced change on a source object; fresh bread anchors that change to the target object. Transferring this relation supplies a visual reference for the unseen pair. We encode the relation with image-text residuals, which measure each seen composition's displacement from its text reference to its visual center.

We realize this idea with \textbf{Shift by Analogy (SBA)}. A seen pair's residual is $r_{ao}=\mu^0_{ao}-t^0_{ao}$, where $t^0_{ao}$ and $\mu^0_{ao}$ are its pretrained text feature and mean normalized image feature. Let $a$ denote sliced, $b$ fresh, $p$ tomato, and $o$ bread. The core \emph{add-subtract-add} construction is
\begin{equation*}
\widetilde r_{ao}=r_{ap}-r_{bp}+r_{bo},\qquad
\widehat\mu_{ao}=t_{ao}+\widetilde r_{ao}.
\end{equation*}
Subtracting $r_{bp}$ from $r_{ap}$ extracts the attribute change while canceling the source object's shared effect; adding $r_{bo}$ supplies the target object's reference state. Figure~\ref{fig:motivation}(b) then follows \emph{shift, then match}. Starting at the target's own text feature $t_{ao}$, the three vectors produce an initial center estimate $\widehat\mu_{ao}$. Refinement yields the visual proxy $q^{\mathrm{cross}}_{ao}$, shown as a diamond. The black star represents the target's visual center formed by encoded images. The shorter proxy-to-center connection depicts improved correspondence for visual matching.

During training, SBA excludes each target composition from its own supports and learns to recognize it using references constructed from other seen pairs. This directly couples reference construction with recognition under target-absent conditions. At inference, the same process gives unseen candidates visual references for matching query images. SBA averages legal residual paths into an analogy library, supplements them with attention over the target object's seen states, and learns refinement jointly with the encoders. Image-conditioned fusion combines proxy matching with semantic and primitive evidence.

The center-cosine comparison in Fig.~\ref{fig:motivation}(b) quantifies this improvement. Over 400 unseen MIT-States compositions, mean cosine with the empirical test-set center rises from \textbf{0.2944} for the text reference to \textbf{0.5395} for SBA's refined cross-object proxy, a gain of \textbf{0.2451}. Residual transfer thus equips unseen compositions with references that align closely with their visual observations. Across UT-Zappos, MIT-States, and C-GQA, SBA consistently outperforms strong compositional prompting and cross-modal baselines under both closed- and open-world evaluation.

Our contributions are threefold:
\begin{itemize}
\item We formulate compositional transfer as constructing visual references for unseen pairs from seen relations. The core add-subtract-add rule operates on observed image-text residuals, making missing compositions visually available through an explicit analogy.
\item We propose SBA, which learns to construct and use references with each target excluded from its supports. An analogy library, same-object context, and joint proxy learning extend this process to unseen compositions.
\item Experiments on three benchmarks demonstrate consistent gains over strong CZSL baselines in closed and open worlds. Visual-center analysis directly verifies improved alignment of constructed references with unseen compositions, supported by ablation studies and visual attribution analysis.
\end{itemize}

%% file: figures/teaser/teaser.tex
\begin{figure}[t]
  \centering
  \captionsetup{skip=4pt}
  \includegraphics[width=\linewidth]{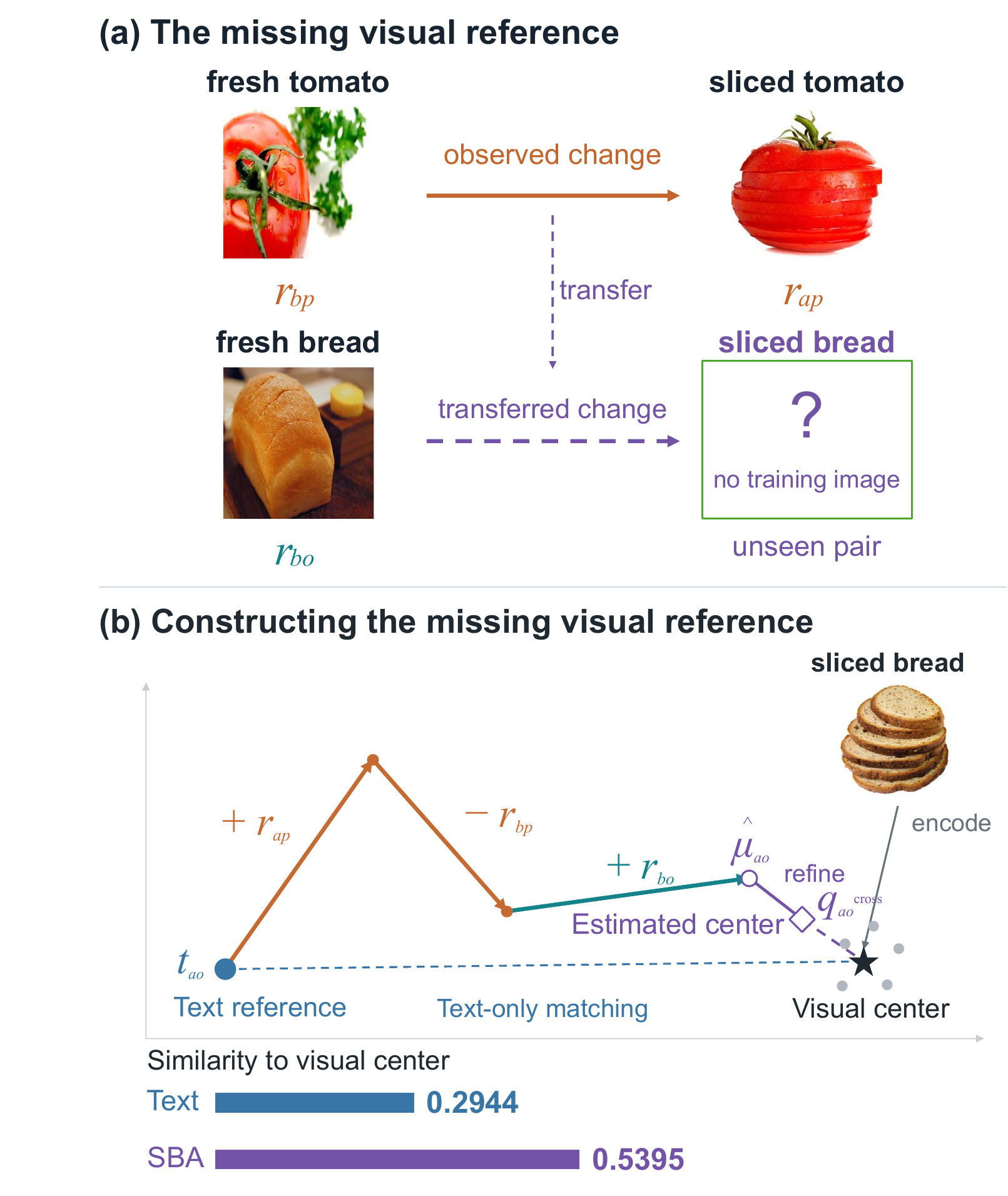}
  \caption{\textbf{Making unseen compositions visually seen.}
  (a) Three seen pairs support the missing sliced bread.
  (b) Constructing the missing visual reference by analogy.
  Cosine similarity to the visual center is shown for text and SBA's cross-object proxy, averaged over 400 unseen MIT-States compositions.}
  \label{fig:motivation}
\end{figure}

%% file: figures/method/method.tex
\begin{figure*}[t]
  \centering
  \captionsetup{skip=4pt}
  \includegraphics[width=\textwidth]{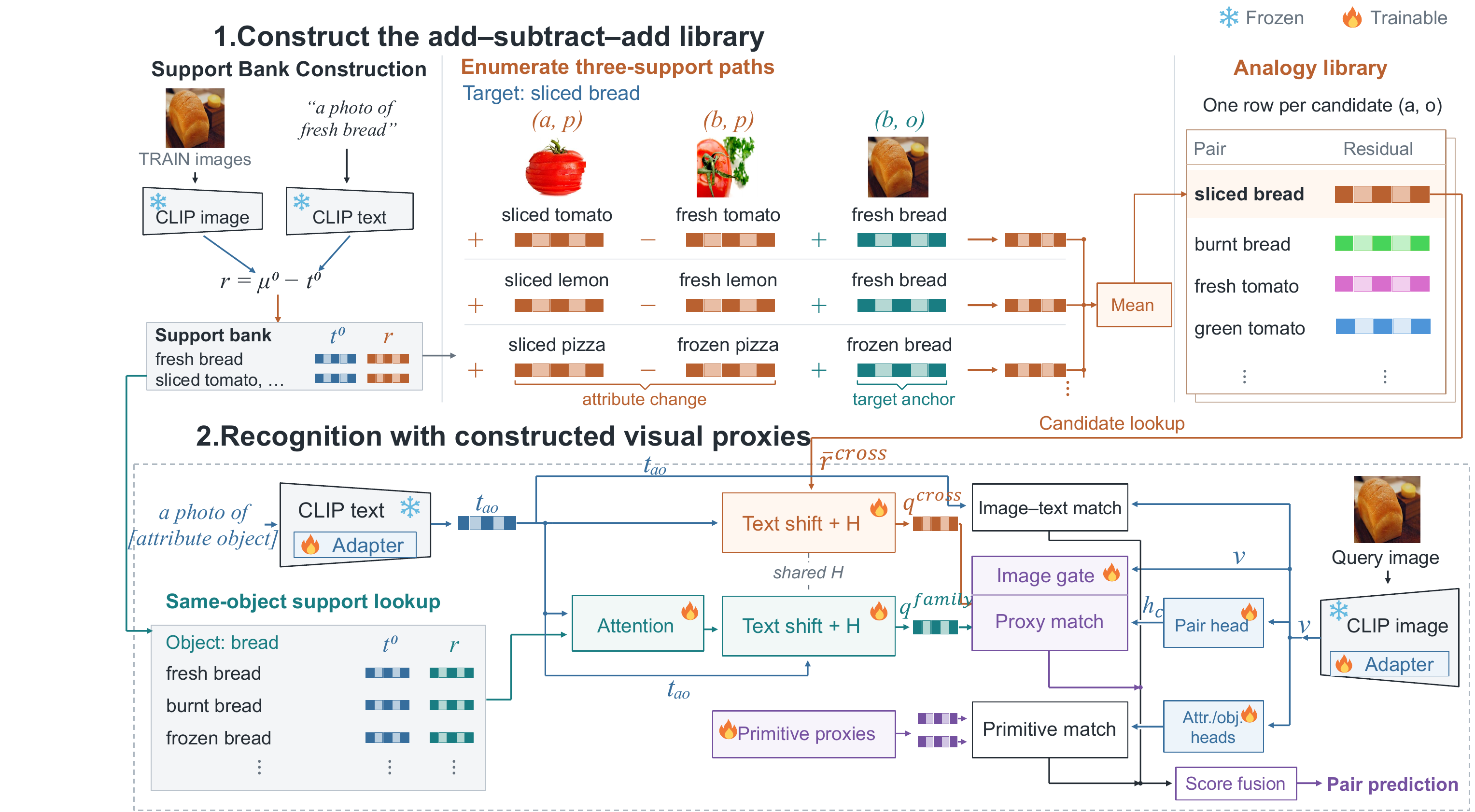}
  \caption{\textbf{Overview of Shift by Analogy (SBA).}
  (1) Seen image-text residuals form a support bank; add-subtract-add paths
  populate the analogy library. (2) Candidate lookup and same-object attention
  construct visual proxies for joint recognition. Flames and snowflakes denote
  trainable and frozen components, respectively.}
  \label{fig:framework}
\end{figure*}

%% file: sections/2_related_work.tex
\section{Related Work}
\label{sec:related}
\paragraph{Compositional representations.}
CZSL learns transferable attribute-object representations from seen combinations. Early approaches compose classifiers~\cite{misra2017red}, attribute operators~\cite{nagarajan2018operators}, and modular networks~\cite{purushwalkam2019modular}; symmetry constraints~\cite{li2020symmetry}, causal modeling~\cite{atzmon2020causal}, and graph embeddings~\cite{naeem2021graph} further structure primitive relations. OADis~\cite{saini2022disentangling} and ADE~\cite{hao2023attention} disentangle visual primitives, while conditional attributes~\cite{wang2023conditional}, contextual specificity~\cite{li2024specificity}, and conditional probability modeling~\cite{wu2025cpf} capture object-dependent appearance. Open-world recognition extends the candidates to the full attribute-object product~\cite{mancini2021open}.

CLIP~\cite{radford2021clip} supports compositional prompting and adaptation~\cite{zhou2022cocoop,chen2022adaptformer}. CSP~\cite{nayak2023csp} and DFSP~\cite{lu2023dfsp} learn primitive prompts; Troika~\cite{huang2024troika} coordinates composition and primitive paths. PLID~\cite{bao2024plid}, LOGICZSL~\cite{wu2025logiczsl}, and EVA~\cite{zhang2026eva} enrich these representations through language distributions, logic, and expert alignment.

\paragraph{Visual proxies and analogical transfer.}
DLM~\cite{hu2024dlm} adapts visual features and composition prototypes to each image; IMAX~\cite{jiang2025imax} models attribute-object interactions in complex space. The image-text modality gap~\cite{liang2022gap} motivates visual references beyond textual prototypes. VP-CMJL~\cite{zhang2025proxy} jointly learns composition and primitive proxies; SAM~\cite{zhang2026sam} combines sparse alignment with unimodal memory, and PCI~\cite{zhu2026pci} uses attribute-conditioned prototype modulation. Visual analogy~\cite{reed2015analogy} transfers changes between contexts, while FlowComposer~\cite{he2026flowcomposer} composes primitive velocity fields for CZSL. SBA constructs candidate visual references from observed residual relations: subtracting two residuals on one object isolates an attribute change, and adding a target-object residual anchors it to its context. Averaging these paths forms the analogy library, complemented by same-object support attention and joint recognition learning.

%% file: sections/3_method.tex
\section{Method}
\label{sec:method}
\textbf{Shift by Analogy (SBA)} realizes the shift-then-match process in Fig.~\ref{fig:motivation}: construct a visual reference for a missing composition, then use it for recognition. Figure~\ref{fig:framework}(1) details library construction. Frozen CLIP features provide seen residuals, and \emph{add-subtract-add} transfers an attribute change to a target-object anchor. Averaging legal paths produces one residual estimate per candidate. Figure~\ref{fig:framework}(2) details recognition: the estimate shifts the candidate's text feature, while attention to the target object's seen states supplies a companion estimate. Shared refinement yields two visual proxies. Training excludes each target from its supports and jointly learns image-gated proxy matching, direct image-text matching, and attribute/object matching. The same construction supplies unseen references at inference. We detail these steps in order.

\subsection{Support Bank Construction}
\label{sec:support-bank}
Let $\mathcal A$ and $\mathcal O$ denote the attribute and object vocabularies, $\mathcal C_s\subseteq\mathcal A\times\mathcal O$ the seen compositions, and $\mathcal D_s=\{(x_i,a_i,o_i)\}_{i=1}^{n}$ the training set. The candidate universe $\mathcal C$ contains the seen pairs and the compositions required for evaluation; $\mathcal C_{\mathrm{test}}\subseteq\mathcal C$ is the allowed test set. We construct candidate supports using $\mathcal D_s$ and the attribute-object relations in $\mathcal C_s$.

Let $f_v^0,f_t^0$ be the pretrained CLIP ViT-L/14 encoders~\cite{radford2021clip}, $N(z)=z/\|z\|_2$, and $\pi(a,o)$ the prompt ``a photo of $a$ $o$''. The embedding dimension is $d=768$. For each seen pair, let $\mathcal I_{ao}=\{i:(a_i,o_i)=(a,o)\}$ and $n_{ao}=|\mathcal I_{ao}|$. We compute its visual center, text reference, and residual in the frozen pretrained space:
\begin{align}
\mu^0_{ao}&={1\over n_{ao}}\sum_{i\in\mathcal I_{ao}}
N(f_v^0(x_i)), \label{eq:visual-center}\\
t^0_{ao}&=N(f_t^0(\pi(a,o))),\qquad
r_{ao}=\mu^0_{ao}-t^0_{ao}. \label{eq:residual}
\end{align}

Each image feature is normalized before averaging, and the visual center retains its magnitude. Thus $r_{ao}$ measures the displacement from a composition's text reference to its observed visual center. We store $\mathcal B=\{(t^0_{ao},r_{ao}):(a,o)\in\mathcal C_s\}$ as the fixed support bank. For target $(a,o)$, we use $\mathcal S_{ao}=\mathcal C_s\setminus\{(a,o)\}$: the target's entire composition is excluded from its supports, while its training images still supervise recognition. This rule applies to both proxy branches and to seen candidates at test time. All stored texts and residuals remain fixed during learning.

\subsection{Add-Subtract-Add Analogy Library}
\label{sec:analogy-library}
For target pair $(a,o)$, we select a source object $p$ and reference attribute $b$ such that $(a,p)$, $(b,p)$, and $(b,o)$ are observed. These three seen pairs define the central construction of SBA:
\begin{equation}
\widetilde r_{ao}^{(b,p)}=r_{ap}-r_{bp}+r_{bo}.
\label{eq:single-analogy}
\end{equation}

Here, $r_{ap}-r_{bp}$ captures the attribute change from $b$ to $a$ on source object $p$, and $r_{bo}$ anchors it to target object $o$. For ``sliced bread'' in Fig.~\ref{fig:framework}(1), this is ``sliced tomato'' minus ``fresh tomato'' plus ``fresh bread''. Other observed states provide further paths, e.g., ``sliced pizza'' minus ``frozen pizza'' plus ``frozen bread''. We average all legal estimates:
\begin{equation}
\begin{split}
\mathcal T_{ao}=\{(b,p):\;&(a,p),(b,p),(b,o)\in\mathcal S_{ao},\\
&b\ne a,\ p\ne o\},
\end{split}
\label{eq:analogy-paths}
\end{equation}
\begin{equation}
\overline r^{\mathrm{cross}}_{ao}
={1\over|\mathcal T_{ao}|}
\sum_{(b,p)\in\mathcal T_{ao}}\widetilde r_{ao}^{(b,p)}.
\label{eq:cross-residual}
\end{equation}

To expose the cancellation, write $r_{ao}=\eta_a+\xi_o+\varepsilon_{ao}$, where $\eta_a$ and $\xi_o$ are attribute and object effects and $\varepsilon_{ao}$ collects their interaction. The three-support construction gives
\begin{equation}
\begin{split}
\widetilde r_{ao}^{(b,p)}
&=\eta_a+\xi_o\\
&\quad+\varepsilon_{ap}-\varepsilon_{bp}+\varepsilon_{bo}.
\end{split}
\label{eq:analogy-effects}
\end{equation}

Subtracting $r_{bp}$ removes the shared source-object effect, and adding $r_{bo}$ cancels the reference-attribute effect while supplying the target-object effect. The resulting estimate retains the desired $\eta_a+\xi_o$ while aggregating transferable interaction context from the supports. We cache $\overline r^{\mathrm{cross}}_{ao}$ and its availability for each candidate to form the \emph{analogy library}. The library stores residual estimates, which will shift each candidate's own text feature during recognition. We also index the same-object supports $\mathcal F_{ao}=\{(b,o)\in\mathcal S_{ao}\}$ in $\mathcal B$ to provide object-specific context for the companion branch.

\subsection{Visual Proxy Construction}
\label{sec:proxy-construction}
We adapt both CLIP encoders with trainable bottleneck paths following Transformer adaptation~\cite{chen2022adaptformer}, while keeping the pretrained weights and support bank fixed. Denoting the adapted encoders by $f_v,f_t$, the current image and candidate text features are
\begin{equation}
v=N(f_v(x)),\qquad t_{ao}=N(f_t(\pi(a,o))),
\label{eq:encoding}
\end{equation}

Each visual and text block uses a width-64 adapter with residual scale 0.1; its parameterization is given in the supplement.

For candidate $(a,o)$, the orange branch in Fig.~\ref{fig:framework}(2) retrieves $\overline r^{\mathrm{cross}}_{ao}$ and shifts the current text feature $t_{ao}$. A correction network $H:\mathbb R^{3d}\to\mathbb R^d$ implements the refinement in Fig.~\ref{fig:motivation}(b), turning the shifted estimate into a visual proxy. With brackets denoting concatenation, the cross-object proxy is
\begin{equation}
\begin{split}
q^{\mathrm{cross}}_{ao}=N\!\bigl(t_{ao}+\overline r^{\mathrm{cross}}_{ao}
+H([t_{ao},\overline r^{\mathrm{cross}}_{ao},0])\bigr).
\end{split}
\label{eq:cross-proxy}
\end{equation}

The teal branch supplies context from the target object's seen states. For ``sliced bread'', these include ``fresh bread'', ``burnt bread'', and ``frozen bread''. The current candidate text queries the fixed texts of $j\in\mathcal F_{ao}$ through semantic similarity and learned query/key projections:
\begin{align}
\ell_{ao,j}&={t_{ao}^{\top}t_j^0\over\tau_f}
+{Q(t_{ao})^{\top}K(t_j^0)\over\sqrt k},
\label{eq:family-logit}\\
w_{ao,j}&={\exp(\ell_{ao,j})\over
\sum_{l\in\mathcal F_{ao}}\exp(\ell_{ao,l})}.
\label{eq:family-weight}
\end{align}

The semantic temperature is $\tau_f=0.07$, and $Q,K:\mathbb R^d\rightarrow\mathbb R^k$ are bias-free linear maps with $k=32$. The first term in Eq.~\eqref{eq:family-logit} supplies a semantic similarity reference; the second learns task-specific support relevance. Using the same weights, we aggregate residuals and support texts:
\begin{equation}
\overline r^{\mathrm{family}}_{ao}=\sum_{j\in\mathcal F_{ao}}w_{ao,j}r_j,
\quad
\overline t^{\mathrm{family}}_{ao}=\sum_{j\in\mathcal F_{ao}}w_{ao,j}t_j^0.
\label{eq:family-aggregation}
\end{equation}

The same-object proxy starts from the current target text and its pooled residual. The discrepancy $t_{ao}-\overline t^{\mathrm{family}}_{ao}$ conditions the correction on the relation between the target and retrieved states:
\begin{equation}
\begin{split}
q^{\mathrm{family}}_{ao}=N\!\bigl(&t_{ao}+\overline r^{\mathrm{family}}_{ao}\\
&+H([t_{ao},\overline r^{\mathrm{family}}_{ao},
t_{ao}-\overline t^{\mathrm{family}}_{ao}])\bigr).
\end{split}
\label{eq:family-proxy}
\end{equation}

Both branches share $H$, a two-layer MLP with hidden width 128, SiLU activation, and a zero-initialized output layer. Their outputs are normalized visual proxies, whereas the library entries remain fixed residuals. When only one branch has supports, its proxy populates both slots; if neither has supports, both default to $t_{ao}$. When both branches are available, each retains its own proxy, ensuring a unified construction across all seen and unseen candidates.

\subsection{Proxy-Guided Recognition}
\label{sec:recognition}
Three independent two-layer residual MLPs $F_u$ project $v$ for composition, attribute, and object matching:
\begin{equation}
h_u=N(v+F_u(v)),\qquad u\in\{c,a,o\}.
\label{eq:visual-projections}
\end{equation}

Following visual proxy learning~\cite{zhang2025proxy}, trainable attribute/object tables are initialized from CLIP texts ``a photo of $a$'' and ``a photo of $o$''. Their normalized rows $p_a,p_o$ supply primitive references. An image-only gate $g(v)=\sigma(G(v))$, initialized to 0.5, balances the two composition proxies and is shared across candidates. The MLP architectures are specified in the supplement.

Let $s$ be CLIP's fixed exponential logit scale. Direct image-text, gated proxy, and primitive matching give
\begin{align}
S_{\mathrm{main}}&=s\,v^\top t_{ao}, \label{eq:main-score}\\
S_{\mathrm{proxy}}&=s\left[g(v)\,h_c^\top q^{\mathrm{cross}}_{ao}\right.
\nonumber\\[-2pt]
&\hspace{13mm}\left.+(1-g(v))\,h_c^\top q^{\mathrm{family}}_{ao}\right],
\label{eq:proxy-score}\\
S_A&=s\,h_a^\top p_a,\qquad S_O=s\,h_o^\top p_o.
\label{eq:primitive-scores}
\end{align}
\begin{equation}
S_{\mathrm{SBA}}=S_{\mathrm{main}}
+\lambda\left(S_{\mathrm{proxy}}+S_A+S_O\right),
\label{eq:sba-score}
\end{equation}

Here $\lambda$ is the fusion weight. We omit $(x,a,o)$ from the scores for readability. The \emph{primitive match} block contributes $S_A+S_O$, while $S_{\mathrm{proxy}}$ connects the add-subtract-add construction and same-object context to each query image. All branches remain active in both training and inference.

\subsection{Joint Learning and Inference}
\label{sec:learning}
For a training label $y=(a,o)$, let $\operatorname{CE}_{\mathcal C_s}(S,y)$ denote cross-entropy over all seen composition scores. Attribute and object cross-entropies use their respective vocabularies. We jointly supervise the original image-text score, the fused prediction, and the proxy branches:
\begin{equation}
\begin{split}
\mathcal L={}&\tfrac12\bigl[\operatorname{CE}_{\mathcal C_s}(S_{\mathrm{main}},y)
+\operatorname{CE}_{\mathcal C_s}(S_{\mathrm{SBA}},y)\bigr]\\
&+\lambda\alpha\bigl[\operatorname{CE}_{\mathcal C_s}(S_{\mathrm{proxy}},y)\\
&\hspace{12mm}+\operatorname{CE}_{\mathcal A}(S_A,a)
+\operatorname{CE}_{\mathcal O}(S_O,o)\bigr].
\end{split}
\label{eq:joint-loss}
\end{equation}

We linearly warm up $\alpha$ from 0 to 1 in the first epoch, then retain $\alpha=1$. All adapters, attention projections, proxy corrections, visual heads, gates, and primitive tables are jointly optimized with SAM/AdamW~\cite{foret2021sam}. CLIP weights, logit scale, and both banks stay fixed.

After training, we cache $t_{ao},q^{\mathrm{cross}}_{ao},q^{\mathrm{family}}_{ao},p_a,p_o$ for all permitted candidates. Each query is encoded once and predicts $\arg\max_{(a,o)\in\mathcal C_{\mathrm{test}}}S_{\mathrm{SBA}}(x,a,o)$ using the same score as in training.

Algorithm~\ref{alg:sba} summarizes support construction, joint optimization, and inference. The analogy library is built once; candidate proxies are recomputed with the current parameters during training and cached after training.

\input{algorithms/sba}

%% file: algorithms/sba.tex
\begin{algorithm}[!t]
\caption{Shift by Analogy: training and inference}
\label{alg:sba}
\footnotesize
\begin{algorithmic}[1]
\Require $\mathcal D_s,\mathcal C_s,\mathcal C,\mathcal C_{\mathrm{test}}$; frozen CLIP encoders.
\Statex \textbf{Support and analogy library construction}
\State Build fixed support bank $\mathcal B$ by Eqs.~\eqref{eq:visual-center}-\eqref{eq:residual}.
\ForAll{$(a,o)\in\mathcal C$}
  \State Exclude $(a,o)$; form $\mathcal S_{ao},\mathcal T_{ao},\mathcal F_{ao}$.
  \State Cache Eq.~\eqref{eq:cross-residual} if $\mathcal T_{ao}\ne\varnothing$, else $0$.
  \State Record availability and the indices of $\mathcal F_{ao}$.
\EndFor
\Statex \textbf{Joint training}
\State Initialize $\Theta$; keep CLIP and both banks fixed.
\ForAll{minibatches across the training epochs}
  \State Encode images and all seen candidate texts by Eq.~\eqref{eq:encoding}.
  \State Retrieve supports and construct both proxies,
  \Statex \hspace{\algorithmicindent} applying the availability rule (Sec.~\ref{sec:proxy-construction}).
  \State Compute projections, gate, and scores by Eqs.~\eqref{eq:visual-projections}-\eqref{eq:sba-score}.
  \State Update $\Theta$ with SAM/AdamW on Eq.~\eqref{eq:joint-loss}.
\EndFor
\Statex \textbf{Inference}
\State Cache final candidate features for $\mathcal C_{\mathrm{test}}$ (Sec.~\ref{sec:learning}).
\ForAll{query images $x$}
  \State Compute $v,h_c,h_a,h_o,g(v)$ once.
  \State Output $\arg\max_{(a,o)\in\mathcal C_{\mathrm{test}}}S_{\mathrm{SBA}}(x,a,o)$.
\EndFor
\end{algorithmic}
\end{algorithm}

%% file: sections/4_experiments.tex
\input{tables/main_comparison}
\section{Experiments}
\label{sec:experiments}
\subsection{Experimental Setup}
\label{sec:experimental-setup}
\textbf{Datasets.} We evaluate on MIT-States~\cite{isola2015states}, UT-Zappos~\cite{yu2014comparisons}, and C-GQA~\cite{naeem2021graph}. They contain 53,753, 29,126, and 39,298 images, respectively. The training/validation/test image counts are 30,338/10,420/12,995 for MIT-States, 22,998/3,214/2,914 for UT-Zappos, and 26,920/7,280/5,098 for C-GQA. The benchmarks cover natural object states, fine-grained footwear, and real-world scenes. Supplementary Sec.~\ref{app:datasets} gives the composition splits, vocabularies, and candidate inventories.

\textbf{Protocols and metrics.} We report both closed- and open-world generalized CZSL. Closed-world candidates contain the seen and held-out compositions; open-world evaluation scores the complete attribute-object product and applies feasibility filtering before seen--unseen calibration. Sweeping the seen-unseen calibration bias yields the area under the curve (AUC) and maximum harmonic mean (HM). S and U are the separate maxima of seen and unseen accuracy. All results are percentages. Accuracies at the maximum-HM operating point are reported in Supplementary Sec.~\ref{app:full-metrics}.

\textbf{Implementation.} SBA uses CLIP ViT-L/14 and $224\times224$ images. We train for 12 epochs with batch size 64, seed 42, and SAM/AdamW. The learning rate warms up from $10^{-5}$ to $10^{-4}$ in epoch 1 and follows cosine decay to $10^{-5}$. The maximum-validation-AUC checkpoint is reused in both test settings. We use $\lambda=0.1$ for C-GQA and MIT-States, and $\lambda=0.445$ for UT-Zappos. Support statistics come from training images, and candidate features are cached after learning. Supplementary Sec.~\ref{app:implementation} specifies all modules and optimization settings.

\subsection{Comparison with Existing Methods}
\label{sec:main-results}
\textbf{Closed-world recognition.} In Table~\ref{tab:main}, SBA achieves the \textbf{highest unseen accuracy} among the compared methods on all three datasets: \textbf{42.8, 55.4, and 77.8} on C-GQA, MIT-States, and UT-Zappos, exceeding the strongest compared baselines by 3.4, 1.5, and 1.5 points. This consistent advantage supports SBA\textquotesingle s core design: \textbf{making unseen compositions visually seen}. Transferring attribute changes between seen compositions and anchoring them to the target object constructs the missing visual reference for an unseen pair. SBA also leads in HM on all three datasets (35.4, 40.9, and 58.0), and in AUC on C-GQA and MIT-States (17.3 and 24.3), with 45.5 AUC on UT-Zappos.

The distinction between U and HM matters for interpreting these gains. The calibration bias applies the same offset to every unseen candidate, preserving the ranking of unseen labels for each image. For a C-GQA image of an unseen composition, even a bias large enough to rule out every seen pair leaves 922 other unseen labels competing with the target. The improvement in U measures discrimination within this set. The improvement in HM shows that SBA can also recognize these targets at a bias that retains accuracy on seen pairs.

\textbf{Comparison with reproduced VP-CMJL.} We reproduce VP-CMJL~\cite{zhang2025proxy} using its official released code and published configurations. SBA improves closed-world AUC by \textbf{2.2, 4.2, and 8.0 points} on C-GQA, MIT-States, and UT-Zappos, with HM gains of 2.6, 3.9, and 8.1 points. Additional comparisons with ClusPro~\cite{qu2025cluspro}, FlowComposer~\cite{he2026flowcomposer}, EVA~\cite{zhang2026eva}, and SAM-CZSL~\cite{zhang2026sam} are provided in Supplementary Sec.~\ref{app:recent}.

\textbf{Open-world recognition.} In the open-world setting, SBA achieves 4.4, 8.3, and 35.4 AUC on C-GQA, MIT-States, and UT-Zappos, surpassing DFSP, PLID, and Troika in both AUC and HM across all three benchmarks, and matching the top 4.4 AUC on C-GQA with the highest seen accuracy (45.9). Against reproduced VP-CMJL, SBA improves AUC/HM by 1.3/1.2 points on MIT-States and 4.9/5.2 points on UT-Zappos. Notably, SBA scales seamlessly to all 278,362 C-GQA candidates by computing candidate proxies once and reusing them across queries.

\subsection{Ablation Study}
\label{sec:component-analysis}
Table~\ref{tab:ablation} evaluates five controlled MIT-States variants under identical initialization, training schedules (four epochs, 1,900 updates), and validation-AUC selection (epoch 3), isolating the two support branches, image gate, primitive supervision, and observed residual input.
\input{tables/ablation}

\textbf{Support construction.} Removing cross-object analogy lowers AUC from 24.3 to 24.2, while removing same-object context drops it to 23.9. Combining both branches integrates transferable attribute shifts with target-object visual experience to achieve the highest AUC. Furthermore, replacing the observed residuals with zeros in $N(t+H([t,0,0]))$ while keeping the correction MLP and all recognition heads lowers AUC from 24.3 to 24.2, confirming that measured residual displacements provide genuine geometric signal beyond learned MLP adaptation.

\textbf{Routing and supervision.} Learned image-conditioned routing outperforms a fixed 0.5 gate in overall AUC (24.3 vs. 24.2, with 40.9 vs. 41.1 HM) by adapting the proxy balance per query. Removing auxiliary attribute/object cross-entropy drops AUC to 23.9; as shown in Table~\ref{tab:primitives}, primitive supervision boosts attribute and object accuracy by 13.4 and 22.5 points across all MIT-States test images, and by 10.0 and 20.5 points on unseen-composition images, supplying stronger primitive evidence that generalizes beyond seen pairs.
\input{tables/primitives}

\subsection{Analysis of Visual References}
\label{sec:reference-analysis}
\textbf{From unseen names to visual references.} The center analysis in Fig.~\ref{fig:motivation} directly tests SBA's motivation. For each of the 400 unseen MIT-States compositions, we average its encoded test images to form an empirical visual center. The macro-averaged cosine increases from 0.2944 for the candidate text to 0.5395 for the refined cross-object proxy; the same-object proxy reaches 0.5522. Thus, the add-subtract-add construction and learned refinement convert a semantic candidate into a reference with substantially stronger visual correspondence.

The center and margin analyses ask different questions about a proxy. We average the center cosines over compositions, so a rare pair has the same weight as a frequent one. The margin is computed per image and compares the score for the correct pair with the highest score among all incorrect pairs. For both measurements the projected image features are held fixed as the reference changes from text to a visual proxy.
\input{figures/experiments/geometry}\input{figures/experiments/sensitivity}\input{figures/experiments/qualitative}\input{figures/experiments/attribution}

\textbf{Separation from competing compositions.} Figure~\ref{fig:geometry} measures the margin between the ground-truth reference cosine and the highest incorrect-reference cosine under the same image feature $h_c$. Cross-object analogy increases the mean margin by 0.00552 on UT-Zappos and boosts strict proxy-only top-1 accuracy from 24.2\% to 28.6\%, while same-object context improves the margin by 0.00250 on MIT-States. Across datasets, the two proxy branches supply distinct geometric evidence that complements semantic and primitive matching in the final prediction.

\textbf{Coverage of missing compositions.} Even with every target excluded from its own supports, cross-object analogy paths cover 95.8\%, 100.0\%, and 99.6\% of unseen pairs on MIT-States, UT-Zappos, and C-GQA, while same-object supports achieve \textbf{100.0\%} coverage on all three benchmarks (including the remaining 17 MIT-States and 4 C-GQA pairs). This complementary coverage guarantees that every unseen composition receives an assembled visual reference from observed pairs.

These coverage counts also specify which evidence enters the score. For a pair with a legal cross-object path, the gate weights estimates built from two different sets of supports. For a pair with only same-object support, the same proxy fills both slots in Eq.~\eqref{eq:proxy-score}, so its contribution is independent of the gate weight. Thus even the 17 MIT-States pairs with no complete analogy path can be scored using visual observations of their target objects.

\subsection{Sensitivity and Efficiency}
\label{sec:sensitivity-efficiency}
\textbf{Fusion weight.} SBA exhibits strong stability across fusion weights $\lambda\in\{0.05,0.1,0.2\}$ (Fig.~\ref{fig:sensitivity}), with AUC varying by only 0.1 points on C-GQA and 0.2 points on MIT-States (both peaking at $\lambda=0.1$). On fine-grained UT-Zappos, a larger proxy weight ($\lambda=0.445$) further raises performance from 44.9 AUC / 57.6 HM (at $\lambda=0.1$) to 45.5 AUC / 58.0 HM.

The fusion weight and the calibration bias play different roles in this comparison. Increasing $\lambda$ gives the gated proxy score and the primitive scores more weight relative to direct image--text matching. Two candidate shoes can share the same object label and still exchange rank as $\lambda$ changes, because their attribute scores and proxy matches differ. The calibration sweep then measures the seen--unseen trade-off for that fixed scoring rule. We use the UT-Zappos weight selected on the validation set in both protocols, which lets us compare the two candidate sets at the same balance of visual and semantic evidence.
\input{tables/efficiency}

\textbf{Computational cost.} SBA introduces only 5.7M, 5.4M, and 6.2M trainable parameters on MIT-States, UT-Zappos, and C-GQA (Table~\ref{tab:efficiency}), with GPU image encoding taking 11.8, 11.7, and 12.0 ms per image-adding negligible (<0.1 ms) overhead over the adapter encoder on MIT-States and UT-Zappos. Because the analogy library, text features, attention outputs, and proxies are cached, inference requires only a single image-encoder pass and lightweight matrix products per query.

\subsection{Qualitative Analysis}
\label{sec:qualitative}
Figure~\ref{fig:qualitative} compares SBA with the independently trained encoder baseline (Base) on C-GQA and MIT-States. On C-GQA, SBA corrects ``pink floor'' to ``tiled floor'', ``large tree'' to ``tall tree'', and ``silver shirt'' to ``white shirt'', while distinguishing ``furry collar'' from ``furry ear'' and recovering ``granite table'' from ``round tray'' across both primitives. On MIT-States, SBA accurately recognizes ``mashed potato'', ``crumpled bag'', ``damp basement'', ``shattered town'', and ``eroded trail'' where the baseline predicts ``whipped potato'', ``empty bag'', ``wet basement'', ``cluttered street'', and ``wide trail''. These examples demonstrate sharper discrimination across surface textures, physical states, and object identities.

In challenging failure cases (Fig.~\ref{fig:qualitative}), SBA often retains one correct primitive-identifying ``package'' in ``red package'' (predicting ``plastic'') and ``muddy'' in ``muddy ground'' (predicting ``lake'')-while visually ambiguous scenes such as ``reflective bowl'' and ``squished bread'' remain difficult for both methods.

\textbf{Visual evidence attribution.} Figure~\ref{fig:attribution} compares Grad-CAM~\cite{selvaraju2017gradcam} maps between SBA and the independently trained encoder baseline (Base). For ``standing zebra'', SBA suppresses background grass hotspots and concentrates attribution on the torso; for ``frayed wire'', responses sharpen around the damaged insulation while suppressing wood-texture distractors; and for ``bent blade'', background activations recede while attribution remains focused along the blade. Two of these corrections preserve the object label: Base predicts ``large zebra'' and ``broken wire.'' For candidates sharing an object, the object score $S_O$ in Eq.~\eqref{eq:sba-score} cancels when comparing their scores. The choice between the two labels therefore depends on the composition and attribute scores. The blade example involves a different decision: changing ``chipped knife'' to ``bent blade'' requires recovering both primitives. In each column, the two maps explain the score for the same ground-truth pair, so their spatial differences can be read against the same target. Supplementary Sec.~\ref{app:attribution} details map extraction and case selection.

%% file: tables/main_comparison.tex
\begin{table*}[t]
\centering
\caption{\textbf{Closed- and open-world results (\%) with CLIP ViT-L/14.} $\dagger$: VP-CMJL reproduced strictly following its official released code and all configurations specified in the paper. Other baselines use published results; dashes denote unreported results. Bold/underline denote best/second best, including ties at the displayed precision. Dataset inventories are in Supplementary Sec.~\ref{app:datasets}.}
\label{tab:main}
\setlength{\tabcolsep}{3.6pt}
\renewcommand{\arraystretch}{0.96}
\resizebox{\textwidth}{!}{%
\begin{tabular}{llrrrr@{\hspace{10pt}}rrrr@{\hspace{10pt}}rrrr}
\toprule
 & & \multicolumn{4}{c}{C-GQA} & \multicolumn{4}{c}{MIT-States} & \multicolumn{4}{c}{UT-Zappos} \\
\cmidrule(lr){3-6}\cmidrule(lr){7-10}\cmidrule(lr){11-14}
Method & Venue & S & U & HM & AUC & S & U & HM & AUC & S & U & HM & AUC \\
\midrule
\multicolumn{14}{c}{\textbf{Closed-world results}} \\
\midrule
CLIP~\cite{radford2021clip} & ICML 2021 & 7.5 & 25.0 & 8.6 & 1.4 & 30.2 & 46.0 & 26.1 & 11.0 & 15.8 & 49.1 & 15.6 & 5.0 \\
CSP~\cite{nayak2023csp} & ICLR 2023 & 28.8 & 26.8 & 20.5 & 6.2 & 46.6 & 49.9 & 36.3 & 19.4 & 64.2 & 66.2 & 46.6 & 33.0 \\
DFSP~\cite{lu2023dfsp} & CVPR 2023 & 38.2 & 32.0 & 27.1 & 10.5 & 46.9 & 52.0 & 37.3 & 20.6 & 66.7 & 71.7 & 47.2 & 36.0 \\
DLM~\cite{hu2024dlm} & AAAI 2024 & 32.4 & 28.5 & 21.9 & 7.3 & 46.3 & 49.8 & 37.4 & 20.0 & 67.1 & 72.5 & 52.0 & 39.6 \\
PLID~\cite{bao2024plid} & ECCV 2024 & 38.8 & 33.0 & 27.9 & 11.0 & 49.7 & 52.4 & 39.0 & 22.1 & 67.3 & 68.8 & 52.4 & 38.7 \\
CDS-CZSL~\cite{li2024specificity} & CVPR 2024 & 38.3 & 34.2 & 28.1 & 11.1 & 50.3 & 52.9 & 39.2 & 22.4 & 63.9 & 74.8 & 52.7 & 39.5 \\
Troika~\cite{huang2024troika} & CVPR 2024 & 41.0 & 35.7 & 29.4 & 12.4 & 49.0 & 53.0 & 39.3 & 22.1 & 66.8 & 73.8 & 54.6 & 41.7 \\
IMAX~\cite{jiang2025imax} & TPAMI 2025 & 39.7 & 35.8 & 29.8 & 12.8 & 48.7 & 53.8 & 39.1 & 21.9 & \underline{69.3} & 70.7 & 54.2 & 40.6 \\
LOGICZSL~\cite{wu2025logiczsl} & CVPR 2025 & 44.4 & \underline{39.4} & \underline{33.3} & \underline{15.3} & \underline{50.8} & \underline{53.9} & \underline{40.5} & \underline{23.4} & \textbf{69.6} & 74.9 & \underline{57.8} & \textbf{45.8} \\
VP-CMJL$^{\dagger}$~\cite{zhang2025proxy} & ICCV 2025 & \underline{45.3} & 38.2 & 32.8 & 15.1 & 45.8 & 52.2 & 37.0 & 20.1 & 63.0 & \underline{76.3} & 49.9 & 37.5 \\
\midrule
\textbf{SBA (Ours)} & & \textbf{45.9} & \textbf{42.8} & \textbf{35.4} & \textbf{17.3} & \textbf{51.4} & \textbf{55.4} & \textbf{40.9} & \textbf{24.3} & 67.7 & \textbf{77.8} & \textbf{58.0} & \underline{45.5} \\
\midrule
\multicolumn{14}{c}{\textbf{Open-world results}} \\
\midrule
CLIP~\cite{radford2021clip} & ICML 2021 & 7.5 & 4.6 & 4.0 & 0.3 & 30.1 & 14.3 & 12.8 & 3.0 & 15.7 & 20.6 & 11.2 & 2.2 \\
CSP~\cite{nayak2023csp} & ICLR 2023 & 28.7 & 5.2 & 6.9 & 1.2 & 46.3 & 15.7 & 17.4 & 5.7 & 64.1 & 44.1 & 38.9 & 22.7 \\
DFSP~\cite{lu2023dfsp} & CVPR 2023 & 38.2 & 7.2 & 10.4 & 2.4 & 47.5 & 18.5 & 19.3 & 6.8 & 66.8 & 60.0 & 44.0 & 30.3 \\
DLM~\cite{hu2024dlm} & AAAI 2024 & --- & --- & --- & --- & --- & --- & --- & --- & --- & --- & --- & --- \\
PLID~\cite{bao2024plid} & ECCV 2024 & 39.1 & 7.5 & 10.6 & 2.5 & 49.1 & 18.7 & 20.4 & 7.3 & 67.6 & 55.5 & 46.6 & 30.8 \\
CDS-CZSL~\cite{li2024specificity} & CVPR 2024 & 37.6 & 8.2 & 11.6 & 2.7 & 49.4 & \textbf{21.8} & \underline{22.1} & \underline{8.5} & 64.7 & 61.3 & 48.2 & 32.3 \\
Troika~\cite{huang2024troika} & CVPR 2024 & 40.8 & 7.9 & 10.9 & 2.7 & 48.8 & 18.4 & 20.1 & 7.2 & 66.4 & 61.2 & 47.8 & 33.0 \\
IMAX~\cite{jiang2025imax} & TPAMI 2025 & 38.7 & 7.9 & 11.2 & 2.5 & 50.2 & 18.6 & 21.4 & 7.6 & \underline{68.4} & 57.3 & 47.5 & 32.3 \\
LOGICZSL~\cite{wu2025logiczsl} & CVPR 2025 & 43.7 & 9.3 & 12.6 & \underline{3.4} & \underline{50.7} & \underline{21.4} & \textbf{22.4} & \textbf{8.7} & \textbf{69.6} & \textbf{63.7} & \textbf{50.8} & \textbf{36.2} \\
VP-CMJL$^{\dagger}$~\cite{zhang2025proxy} & ICCV 2025 & \underline{45.3} & \textbf{11.3} & \textbf{14.7} & \textbf{4.4} & 45.8 & 19.4 & 20.0 & 7.0 & 62.9 & \underline{63.3} & 45.2 & 30.5 \\
\midrule
\textbf{SBA (Ours)} & & \textbf{45.9} & \underline{11.0} & \underline{14.3} & \textbf{4.4} & \textbf{51.4} & 20.7 & 21.2 & 8.3 & 67.7 & 61.7 & \underline{50.4} & \underline{35.4} \\
\bottomrule
\end{tabular}}
\end{table*}

%% file: tables/ablation.tex
\begin{table}[t]
\centering
\caption{\textbf{MIT-States training ablations.} All configurations use matched four-epoch training and validation selection; $\Delta$ is variant minus SBA AUC, computed before rounding.}
\label{tab:ablation}
\footnotesize
\setlength{\tabcolsep}{3.1pt}
\renewcommand{\arraystretch}{1.07}
\begin{tabular}{lrrr}
\toprule
Configuration & HM & AUC & $\Delta$AUC \\
\midrule
w/o cross-object & 40.9 & 24.2 & -0.1 \\
w/o same-object & 40.8 & 23.9 & -0.4 \\
Uniform gate & \textbf{41.1} & 24.2 & -0.1 \\
w/o primitive loss & 40.7 & 23.9 & -0.4 \\
w/o residual input & 40.9 & 24.2 & 0.0 \\
\midrule
\textbf{SBA} & 40.9 & \textbf{24.3} & --- \\
\bottomrule
\end{tabular}
\end{table}

%% file: tables/primitives.tex
\begin{table}[t]
\centering
\caption{\textbf{Primitive recognition on MIT-States.} Accuracy over all attributes or objects; unseen refers to unseen compositions.}
\label{tab:primitives}
\footnotesize
\setlength{\tabcolsep}{3.1pt}
\renewcommand{\arraystretch}{1.07}
\begin{tabular}{lrrrr}
\toprule
 & \multicolumn{2}{c}{All images} & \multicolumn{2}{c}{Unseen images} \\ \cmidrule(lr){2-3}\cmidrule(lr){4-5} Model & Attr. & Obj. & Attr. & Obj. \\
\midrule
w/o primitive loss & 17.7 & 28.1 & 17.1 & 27.1 \\
\textbf{SBA} & \textbf{31.1} & \textbf{50.6} & \textbf{27.1} & \textbf{47.6} \\
\bottomrule
\end{tabular}
\end{table}

%% file: figures/experiments/geometry.tex
\begin{figure}[t]
\centering
\includegraphics[width=\columnwidth]{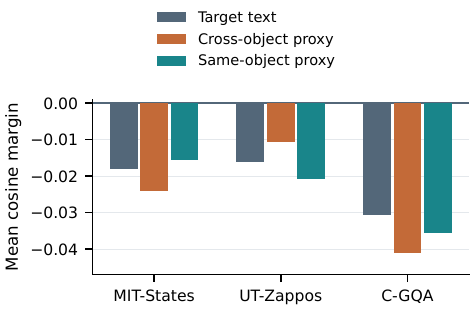}
\caption{\textbf{Separation from the hardest negative.} Mean correct-minus-incorrect cosine on unseen images, using the same $h_c$. Higher is better.}
\label{fig:geometry}
\end{figure}

%% file: figures/experiments/sensitivity.tex
\begin{figure}[t]
\centering
\includegraphics[width=\columnwidth]{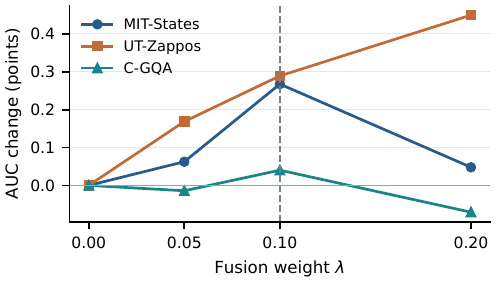}
\caption{\textbf{Fusion-weight sensitivity.} AUC changes relative to $\lambda=0$ on the selected checkpoints for $\lambda\in\{0,0.05,0.1,0.2\}$. The dashed line marks $\lambda=0.1$.}
\label{fig:sensitivity}
\end{figure}

%% file: figures/experiments/qualitative.tex
\begin{figure*}[t]
\centering
\includegraphics[width=\textwidth]{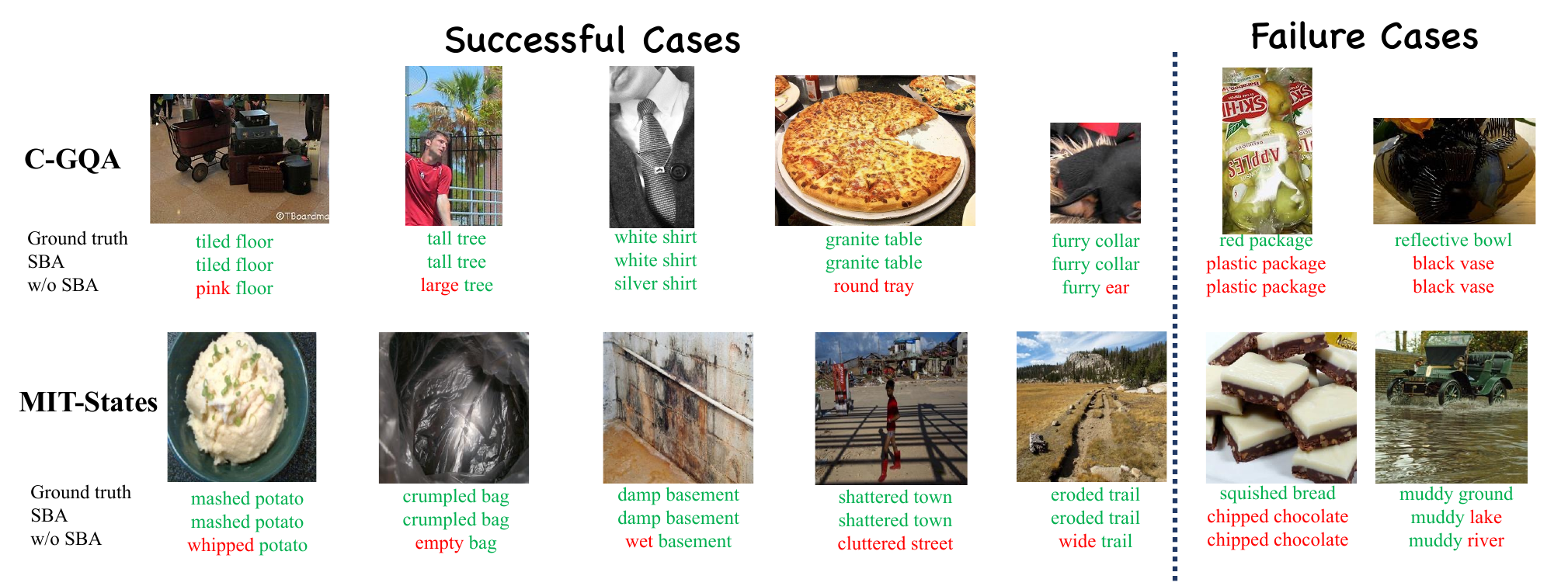}
\caption{\textbf{Qualitative comparison on C-GQA and MIT-States.} Successful and failure cases; green/red denotes correct/incorrect labels.}
\label{fig:qualitative}
\end{figure*}

%% file: figures/experiments/attribution.tex
\begin{figure}[t]
\centering
\includegraphics[width=\columnwidth]{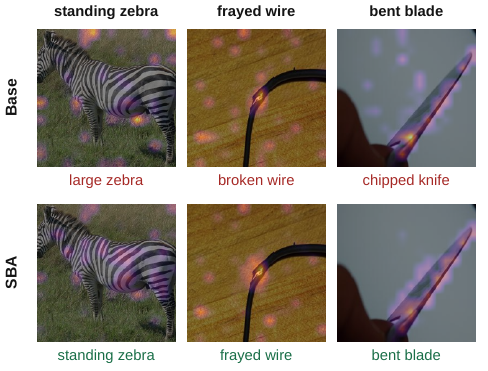}
\caption{\textbf{Visual evidence attribution.} Ground-truth labels appear above each column; predictions are below each map. Both Grad-CAM maps explain the same ground-truth composition score with a shared color scale within each pair.}
\label{fig:attribution}
\end{figure}

%% file: tables/efficiency.tex
\begin{table}[t]
\centering
\caption{\textbf{Trainable size and GPU image encoding.} Times are milliseconds per image. The encoder control uses the same adapters; candidate scoring is separate.}
\label{tab:efficiency}
\footnotesize
\setlength{\tabcolsep}{3.1pt}
\renewcommand{\arraystretch}{1.07}
\begin{tabular}{lrrr}
\toprule
Dataset & Parameters & SBA (ms) & Encoder (ms) \\
\midrule
C-GQA & 6.2M & 12.0 & --- \\
MIT-States & 5.7M & 11.8 & 11.7 \\
UT-Zappos & 5.4M & 11.7 & 11.6 \\
\bottomrule
\end{tabular}
\end{table}

%% file: sections/5_conclusion.tex
\section{Conclusion}
\label{sec:conclusion}
Shift by Analogy constructs a visual reference for an attribute--object pair before seeing any training image of that pair. Its add-subtract-add rule transfers an observed attribute change to the target object, and same-object attention supplies a reference wherever a cross-object path is missing. Together the two branches cover every unseen test composition on all three benchmarks using only supports drawn from observed training pairs. On MIT-States, the refined cross-object proxy improves mean cosine similarity to the unseen visual center by 0.2451 over text. In the final recognition model, these proxies complement semantic and primitive scores, yielding the highest closed-world unseen accuracy and HM across the three datasets in Table~\ref{tab:main}.

%% file: tables/supplement/complete/comparison.tex
\centering
\caption{\textbf{Additional comparisons (\%).} Within each protocol and column, bold marks the best result and underlining marks the second best. Ties share a rank at the displayed precision.}
\label{tab:s-comparison}
\small\setlength{\tabcolsep}{4pt}\renewcommand{\arraystretch}{1.10}
\begingroup\setbox0=\hbox{\begin{tabular*}{\textwidth}{@{\extracolsep{\fill}}lrrrr@{\hspace{10pt}}rrrr@{\hspace{10pt}}rrrr@{}}
\toprule
 & \multicolumn{4}{c}{C-GQA} & \multicolumn{4}{c}{MIT-States} & \multicolumn{4}{c}{UT-Zappos} \\
\cmidrule(lr){2-5}\cmidrule(lr){6-9}\cmidrule(lr){10-13}
Method & S & U & HM & AUC & S & U & HM & AUC & S & U & HM & AUC \\
\midrule
\multicolumn{13}{l}{\itshape Closed world} \\
ClusPro~\cite{qu2025cluspro} & \underline{45.6} & 38.2 & \underline{33.9} & \underline{15.5} & 48.7 & 51.3 & 38.2 & 21.1 & \textbf{69.4} & \underline{75.4} & \underline{57.6} & \underline{45.3} \\
FlowComposer~\cite{he2026flowcomposer} & 44.6 & \underline{38.5} & 33.1 & 14.9 & \underline{49.1} & \underline{52.3} & \underline{38.4} & \underline{21.7} & 63.4 & 71.4 & 49.2 & 36.7 \\
EVA~\cite{zhang2026eva} & 42.8 & 32.0 & 29.1 & 11.9 & 45.3 & 48.3 & 35.2 & 18.2 & 64.3 & 68.1 & 49.5 & 35.5 \\
SAM-CZSL~\cite{zhang2026sam} & \underline{45.6} & 35.6 & 29.8 & 13.6 & 48.2 & 51.4 & 37.2 & 20.4 & 66.4 & 69.2 & 50.9 & 37.3 \\
\textbf{SBA} & \textbf{45.9} & \textbf{42.8} & \textbf{35.4} & \textbf{17.3} & \textbf{51.4} & \textbf{55.4} & \textbf{40.9} & \textbf{24.3} & \underline{67.7} & \textbf{77.8} & \textbf{58.0} & \textbf{45.5} \\
\midrule
\multicolumn{13}{l}{\itshape Open world} \\
ClusPro~\cite{qu2025cluspro} & \underline{45.6} & \underline{10.3} & \underline{14.1} & \underline{4.1} & 48.7 & 19.6 & 20.7 & 7.6 & \textbf{69.4} & \underline{47.6} & \underline{42.9} & \underline{27.1} \\
FlowComposer~\cite{he2026flowcomposer} & 44.5 & 8.7 & 11.8 & 3.2 & \underline{49.1} & \underline{20.7} & \underline{21.1} & \underline{7.9} & 63.4 & 43.3 & 37.1 & 21.3 \\
EVA~\cite{zhang2026eva} & 42.2 & 8.9 & 12.8 & 3.4 & 44.2 & 15.8 & 16.5 & 5.3 & 64.3 & 46.5 & 40.2 & 24.3 \\
SAM-CZSL~\cite{zhang2026sam} & 45.3 & 10.1 & 13.1 & 3.9 & 46.4 & \textbf{21.3} & 21.0 & 7.8 & 66.5 & 40.1 & 39.2 & 22.9 \\
\textbf{SBA} & \textbf{45.9} & \textbf{11.0} & \textbf{14.3} & \textbf{4.4} & \textbf{51.4} & \underline{20.7} & \textbf{21.2} & \textbf{8.3} & \underline{67.7} & \textbf{61.7} & \textbf{50.4} & \textbf{35.4} \\
\bottomrule\end{tabular*}}
\noindent\makebox[\textwidth][c]{\ifdim\wd0>\textwidth\resizebox{\textwidth}{!}{\box0}\else\box0\fi}\par\endgroup

%% file: sections/6_appendix.tex
We reproduced ClusPro~\cite{qu2025cluspro}, FlowComposer~\cite{he2026flowcomposer}, EVA~\cite{zhang2026eva}, and SAM-CZSL~\cite{zhang2026sam} following the methods described in their respective papers. Table~\ref{tab:s-comparison} reports our measurements on the three benchmarks under both protocols.

\paragraph{Recognition of unseen compositions.}
Compared with the strongest reproduction on each dataset, SBA gains 4.4 points in closed-world unseen accuracy on C-GQA (over FlowComposer), 3.1 on MIT-States (over FlowComposer), and 2.4 on UT-Zappos (over ClusPro). The largest gain occurs on C-GQA, where the model must distinguish an unseen pair from 6,514 other candidate compositions. SBA also achieves the highest closed-world HM on all three datasets (35.4, 40.9, and 58.0, respectively).

\paragraph{Open-world evaluation.}
The open-world protocol applies feasibility filtering to candidate attribute--object pairs before seen--unseen calibration.

\section{Dataset Splits and Candidate Inventories}
\label{app:datasets}
\input{tables/supplement/complete/datasets}
On C-GQA, SBA uses 413 attributes and 674 objects (Table~\ref{tab:s-datasets}). The 5,592 training pairs and 923 unseen test pairs give 6,515 closed-world test candidates. The full attribute--object vocabulary contains $413\times674=278{,}362$ pairs.

\section{Balanced Recognition Operating Points}
\label{app:full-metrics}
\input{tables/supplement/complete/operating}
Table~\ref{tab:s-operating} gives seen and unseen accuracy at one shared calibration bias. For example, the UT-Zappos closed-world result pairs 58.2\% seen with 57.9\% unseen accuracy at maximum HM.

\paragraph{UT-Zappos fusion weight.}
We selected the fusion weight $\lambda=0.445$ on the UT-Zappos validation set and used it for both closed-world and open-world evaluation.

\section{Implementation Details}\label{app:implementation}
\paragraph{Trainable modules.}
Each of the 24 visual and 12 text transformer blocks adds a width-64 adapter in parallel with its feed-forward layer. For layer-normalized input $z$ of width $D$, the update is
\begin{equation}
\begin{split}
A(z)={}&\frac{\|z\|_2}{\sqrt{64}}W_{\rm up}\\[-2pt]
&\operatorname{GELU}\!\left(W_{\rm down}\frac{\sqrt D\,z}{\|z\|_2}+b_{\rm down}\right).
\end{split}
\end{equation}
Here $D=1024$ for vision and $D=768$ for text. Each block adds $0.1A(z)$. We initialize $W_{\rm down}$ from $\mathcal{N}(0,1/D)$, and the other adapter parameters start at zero. The shared proxy correction has widths $2304\to128\to768$. Each of the three visual heads uses a $768\to128\to768$ residual projection with a SiLU nonlinearity. Their output layers and that of the proxy correction start at zero. The gate uses a width-32 hidden layer and starts at 0.5. Same-object attention uses width-32 queries and keys and semantic temperature 0.07. We initialize each primitive proxy by encoding its name with the frozen text encoder.

\paragraph{Optimization and caching.}
We retain FP32 precision and freeze CLIP's logit scale. SAM uses radius 0.05. AdamW uses $(\beta_1,\beta_2)=(0.9,0.999)$, $\epsilon=10^{-8}$, and no weight decay. We clip the gradient norm to 5 after the second backward pass. In the first epoch the auxiliary loss multiplier increases from zero to one over minibatches and stays at one thereafter. We retain the last partial batch. Candidate text vectors and proxies are cached once, while image features and the gate are evaluated per query.
\input{tables/supplement/complete/cost}
Table~\ref{tab:s-cost} separates the one-time text encoding cost for the closed-world candidate cache from the recurring optimization cost. The memory measurement covers PyTorch's reserved allocation during training.

\section{Visual Attribution Protocol}\label{app:attribution}
For each map we differentiate the score of the same ground-truth composition at the first layer normalization of the final visual block. We average gradients over the $16\times16$ patch grid to obtain channel weights, take the rectified weighted sum of activations, and normalize the positive map to unit mass. We overlay the map on the $224\times224$ model input by bilinear interpolation at 45\% opacity. Both maps for each image use the same color scale.

For the examples in the main paper, we considered 253 corrected predictions on MIT-States and 34 on C-GQA at the fixed evaluation biases. We ranked the pairs by total variation between their normalized attribution maps and inspected the images for interpretability.

\section{Unseen Visual Centers Across Datasets}
For an unseen pair $c$, we average its unit-normalized projected test features $h_c(x)$ and normalize the mean to obtain an empirical direction $\mu_c$. For proxy $q_c$ and candidate text $t_c$, we compute
\begin{equation}
\Delta_c = \cos(q_c,\mu_c)-\cos(t_c,\mu_c).
\end{equation}
Each composition receives equal weight. We evaluate the two proxies separately because the gate mixes them differently for each image. We group test features by their ground-truth composition.
\input{tables/supplement/complete/centers}
\paragraph{Effect of sample size.} Both proxies move closer to the correct visual center on all three datasets (Table~\ref{tab:s-centers}). The gain is largest on UT-Zappos and remains positive on MIT-States and C-GQA when each center must contain at least five images. On C-GQA, 12 of 923 unseen pairs meet the five-image threshold.

\paragraph{Split-half repeatability.} For the retained pairs, we also randomly split each composition's images in half and compute two centers. Over 20 splits, their mean cosine similarity (scaled by 100) is 92.6 on C-GQA, 96.4 on MIT-States, and 99.2 on UT-Zappos.

%% file: tables/supplement/complete/datasets.tex
\begin{table}[H]
\centering
\caption{Composition inventories used by SBA. Closed-world candidates are the union of training pairs and the held-out pairs for the indicated split.}
\label{tab:s-datasets}
\footnotesize\setlength{\tabcolsep}{4pt}\renewcommand{\arraystretch}{1.10}
\begingroup\setbox0=\hbox{\begin{tabular}{lrrr}
\toprule
Inventory & C-GQA & MIT-States & UT-Zappos \\
\midrule
Attributes & 413 & 115 & 16 \\
Objects & 674 & 245 & 12 \\
Training pairs & 5,592 & 1,262 & 83 \\
Unseen validation pairs & 1,252 & 300 & 15 \\
Unseen test pairs & 923 & 400 & 18 \\
Closed validation & 6,844 & 1,562 & 98 \\
Closed test & 6,515 & 1,662 & 101 \\
Open vocabulary product & 278,362 & 28,175 & 192 \\
\bottomrule\end{tabular}}
\noindent\makebox[\columnwidth][c]{\ifdim\wd0>\columnwidth\resizebox{\columnwidth}{!}{\box0}\else\box0\fi}\par\endgroup
\end{table}

%% file: tables/supplement/complete/operating.tex
\begin{table}[H]
\centering
\caption{SBA seen and unseen accuracy (\%) at the bias that maximizes HM for each protocol.}
\label{tab:s-operating}
\footnotesize\setlength{\tabcolsep}{4pt}\renewcommand{\arraystretch}{1.10}
\begingroup\setbox0=\hbox{\begin{tabular}{lrrrr}
\toprule
 & \multicolumn{2}{c}{Closed} & \multicolumn{2}{c}{Open} \\
\cmidrule(lr){2-3}\cmidrule(lr){4-5}
Dataset & $S^*$ & $U^*$ & $S^*$ & $U^*$ \\
\midrule
C-GQA & 37.5 & 33.6 & 29.4 & 9.4 \\
MIT-States & 38.2 & 44.2 & 29.3 & 16.6 \\
UT-Zappos & 58.2 & 57.9 & 55.7 & 46.0 \\
\bottomrule\end{tabular}}
\noindent\makebox[\columnwidth][c]{\ifdim\wd0>\columnwidth\resizebox{\columnwidth}{!}{\box0}\else\box0\fi}\par\endgroup
\end{table}

%% file: tables/supplement/complete/cost.tex
\begin{table}[H]
\centering
\caption{Training and candidate-encoding costs. Update time includes both SAM passes. Peak reserved GPU memory is measured over 20 full-batch updates.}
\label{tab:s-cost}
\footnotesize\setlength{\tabcolsep}{4pt}\renewcommand{\arraystretch}{1.10}
\begingroup\setbox0=\hbox{\begin{tabular}{lrrr}
\toprule
Measurement & C-GQA & MIT-States & UT-Zappos \\
\midrule
Training (h) & 33.4 & 11.5 & 4.6 \\
Update (s) & 23.4 & 6.9 & 3.7 \\
Text cache (s) & 4.0 & 1.1 & 0.2 \\
Reserved (GiB) & 29.9 & 40.0 & 29.1 \\
\bottomrule\end{tabular}}
\noindent\makebox[\columnwidth][c]{\ifdim\wd0>\columnwidth\resizebox{\columnwidth}{!}{\box0}\else\box0\fi}\par\endgroup
\end{table}

%% file: tables/supplement/complete/centers.tex
\begin{table}[H]
\centering
\caption{Gain in cosine similarity to the empirical unseen visual center relative to text (cosine units $\times 100$). The last two columns restrict the calculation to compositions with at least five test images.}
\label{tab:s-centers}
\footnotesize\setlength{\tabcolsep}{4pt}\renewcommand{\arraystretch}{1.10}
\begingroup\setbox0=\hbox{\begin{tabular}{llrrr}
\toprule
Dataset & Proxy & All & $n\!\geq\!5$ & Pairs ($n\!\geq\!5$) \\
\midrule
C-GQA & Cross-object & 1.6 & 2.8 & 12 \\
 & Same-object & 2.4 & 3.7 & 12 \\
MIT-States & Cross-object & 24.5 & 24.8 & 362 \\
 & Same-object & 25.8 & 26.0 & 362 \\
UT-Zappos & Cross-object & 57.6 & 57.6 & 18 \\
 & Same-object & 56.4 & 56.4 & 18 \\
\bottomrule\end{tabular}}
\noindent\makebox[\columnwidth][c]{\ifdim\wd0>\columnwidth\resizebox{\columnwidth}{!}{\box0}\else\box0\fi}\par\endgroup
\end{table}